\documentclass{article}

\usepackage[utf8]{inputenc}
\usepackage{multirow}
\usepackage{tabularx}
\usepackage[table]{xcolor}
\usepackage{siunitx}
\usepackage{caption}
\usepackage{microtype}
\usepackage{graphicx}
\usepackage{subcaption}
\usepackage{booktabs} 
\definecolor{amemgray}{gray}{0.92}

\usepackage{csquotes}
\usepackage[table]{xcolor} 
\usepackage{multirow}
\usepackage{tabularx}

\usepackage{tcolorbox}
\usepackage{fontawesome5}
\usepackage{xcolor}
\usepackage{listings}

\definecolor{promptbg}{HTML}{F8F9FA}
\definecolor{promptframe}{HTML}{4A90D9}
\definecolor{prompttext}{HTML}{2D3748}
\definecolor{accentblue}{HTML}{3182CE}
\definecolor{accentgreen}{HTML}{38A169}
\definecolor{accentyellow}{HTML}{D69E2E}
\definecolor{accentpurple}{HTML}{805AD5}
\definecolor{accentred}{HTML}{F38BA8}
\definecolor{accentorange}{HTML}{FAB387}
\definecolor{accentteal}{HTML}{94E2D5}
\definecolor{codebg}{HTML}{313244}
\definecolor{commentcolor}{HTML}{6C7086}

\definecolor{accentred}{HTML}{E53E3E}
\definecolor{codebg}{HTML}{EDF2F7}
\definecolor{commentcolor}{HTML}{718096}
\tcbuselibrary{skins,breakable}
\newtcolorbox{systemprompt}[1][]{
  enhanced,
  breakable,
  colback=promptbg,
  colframe=promptframe,
  colbacktitle=promptframe,
  coltitle=promptbg,
  fonttitle=\bfseries\sffamily,
  title={\faRobot\hspace{0.5em}#1},
  arc=4pt,
  boxrule=1.5pt,
  left=10pt, right=10pt, top=8pt, bottom=8pt,
  toptitle=4pt, bottomtitle=4pt,
  shadow={2pt}{-2pt}{0pt}{black!50},
  fontupper=\small\ttfamily\color{prompttext},
}

\newcommand{\tool}[1]{\textcolor{accentgreen}{\textbf{#1}}}
\newcommand{\keyword}[1]{\textcolor{accentblue}{\textbf{#1}}}

\newcommand{\important}[1]{\textcolor{accentred}{\textbf{#1}}}
\newcommand{\stage}[1]{\textbf{#1}}

\renewcommand{\section}[1]{%
  \textcolor{accentpurple}{\textbf{\faChevronRight\hspace{0.3em}#1}}%
}

\newcommand{\correct}[1]{\textcolor{accentgreen}{\textbf{#1}}}
\newcommand{\wrong}[1]{\textcolor{accentred}{\textbf{#1}}}
\newcommand{\variable}[1]{\textcolor{accentteal}{\{#1\}}}
\usepackage{listings}
\usepackage{inconsolata} 
\usepackage{pifont}    
\usepackage{enumitem}  
\setlist[itemize,1]{leftmargin=1.2em,itemsep=2pt,topsep=2pt}

\usepackage{hyperref}

\usepackage[preprint]{icml2026}

\usepackage{amsmath}
\usepackage{amssymb}
\usepackage{mathtools}
\usepackage{amsthm}

\usepackage[capitalize,noabbrev]{cleveref}

\theoremstyle{plain}

\theoremstyle{definition}

\theoremstyle{remark}

\usepackage[textsize=tiny]{todonotes}

\definecolor{convblue}{HTML}{0B6E99}
\definecolor{qagreen}{HTML}{2F7D32}
\definecolor{softgray}{HTML}{F5F6F7}
\definecolor{darkgray}{HTML}{333333}

\lstdefinestyle{jsonstyle}{
  basicstyle=\ttfamily\small,
  backgroundcolor=\color{softgray},
  frame=single,
  rulecolor=\color{darkgray},
  columns=fullflexible,
  breaklines=true,
  showstringspaces=false,
  literate=
   *{:}{{\textcolor{darkgray}{:}}}{1}
    {,}{{\textcolor{darkgray}{,}}}{1}
    {[}{{\textcolor{darkgray}{[}}}{1}
    {]}{{\textcolor{darkgray}{]}}}{1}
    {\{}{{\textcolor{darkgray}{\{}}}{1}
    {\}}{{\textcolor{darkgray}{\}}}}{1}
}

\icmltitlerunning{M2A: Multimodal Memory Agent for Personalized Interactions}

\begin{document}

\twocolumn[
  \icmltitle{M$^{2}$A: Multimodal Memory Agent with \\
    Dual-Layer Hybrid Memory for Long-Term Personalized Interactions}

  \icmlsetsymbol{equal}{*}
  \icmlsetsymbol{leader}{†}
  \icmlsetsymbol{corr}{$\dagger$}

  \begin{icmlauthorlist}
    \icmlauthor{Junyu Feng}{xjtu,equal}
    \icmlauthor{Binxiao Xu}{xjtu,pku,equal} 
    \icmlauthor{Jiayi Chen}{cuhk}
    \icmlauthor{Mengyu Dai}{nankai}
    \icmlauthor{Cenyang Wu}{pku}
    \icmlauthor{Haodong Li}{south}
    \icmlauthor{Bohan Zeng}{pku} \\
    \icmlauthor{Yunliu Xie}{pku,leader}
    \icmlauthor{Hao Liang}{pku}
    \icmlauthor{Ming Lu}{intel}
    \icmlauthor{Wentao Zhang}{pku,corr}
 
  \end{icmlauthorlist}

  \icmlaffiliation{pku}{Peking University, Beijing, China}
  \icmlaffiliation{xjtu}{Xi'an Jiaotong University, Xi'an, China}
  \icmlaffiliation{cuhk}{The Chinese University of Hong Kong, Hong Kong, China}
  \icmlaffiliation{nankai}{Nankai University, tianjin, China}
  \icmlaffiliation{south}{South China University of Technology, Guangzhou, China}
  \icmlaffiliation{intel}{Intel, Beijing, China}
  
  \icmlcorrespondingauthor{Wentao Zhang}{wentao.zhang@pku.edu.cn}

  \icmlkeywords{Machine Learning, ICML}

  \vskip 0.3in
]


\printAffiliationsAndNotice{
  \icmlEqualContribution
 \textsuperscript{†}~Project~leader
  \textsuperscript{\textdagger}~Corresponding~author
}

\begin{abstract}
This work addresses the challenge of personalized question answering in long-term human-machine interactions: when conversational history spans weeks or months and exceeds the context window, existing personalization mechanisms struggle to continuously absorb and leverage users' incremental concepts, aliases, and preferences. Current personalized multimodal models are predominantly static—concepts are fixed at initialization and cannot evolve during interactions. We propose \textbf{M$^{2}$A}, an agentic dual-layer hybrid memory system that maintains personalized multimodal information through online updates. The system employs two collaborative agents: ChatAgent manages user interactions and autonomously decides when to query or update memory, while MemoryManager breaks down memory requests from ChatAgent into detailed operations on the dual-layer memory bank, which couples a RawMessageStore (immutable conversation log) with a SemanticMemoryStore (high-level observations), providing memories at different granularities. In addition, we develop a reusable data synthesis pipeline that injects concept-grounded sessions from Yo'LLaVA and MC-LLaVA into LoCoMo long conversations while preserving temporal coherence. Experiments show that M$^{2}$A significantly outperforms baselines, demonstrating that transforming personalization from one-shot configuration to a co-evolving memory mechanism provides a viable path for high-quality individualized responses in long-term multimodal interactions. The code is available at \url{https://github.com/Little-Fridge/M2A}.
\end{abstract}

\section{Introduction}
\label{sec:intro}

Large vision-language models (VLMs) have achieved strong performance on multimodal instruction following, visual question answering, and open-ended dialogue~\cite{li2024llava_onevision,internvl2-2024,qwen2-vl-2024,llava-interleave-2024}. However, these models are primarily trained for generic, ``anonymous'' users and lack mechanisms to explicitly capture individual concepts, naming conventions, or stylistic preferences. Consequently, their responses are often broadly acceptable but insufficiently personalized~\cite{hao2025rap}.

Existing personalization methods fall into two categories. The first internalizes user-provided concept images into model representations, enabling recognition of personalized visual entities within LLaVA-like architectures (e.g., Yo'LLaVA, MC-LLaVA)~\cite{nguyen2024yollava,an2024mcllava}. The second adopts retrieval-augmented paradigms (e.g., RAP), storing user-related information externally and dynamically retrieving it at inference time for better scalability~\cite{hao2025rap}. Despite their differences, most approaches assume personalization is static. In practice, personalization is incremental: users continually refine concepts with new attributes, aliases, and preferences, which static systems cannot effectively absorb.

Meanwhile, long-term conversations quickly exceed context windows, requiring external memory and selective retrieval~\cite{maharana2024locomo,memgpt-2023,zhong2024memorybank, luo2024llm}. However, existing memory systems largely focus on text, offering limited support for multimodal concepts, fine-grained updates, or editable memory structures specifically designed for human-machine interactions. Addressing these limitations requires a unified framework for editable multimodal personalization.

\begin{figure*}[t]
   \centering
   \includegraphics[width=\linewidth]{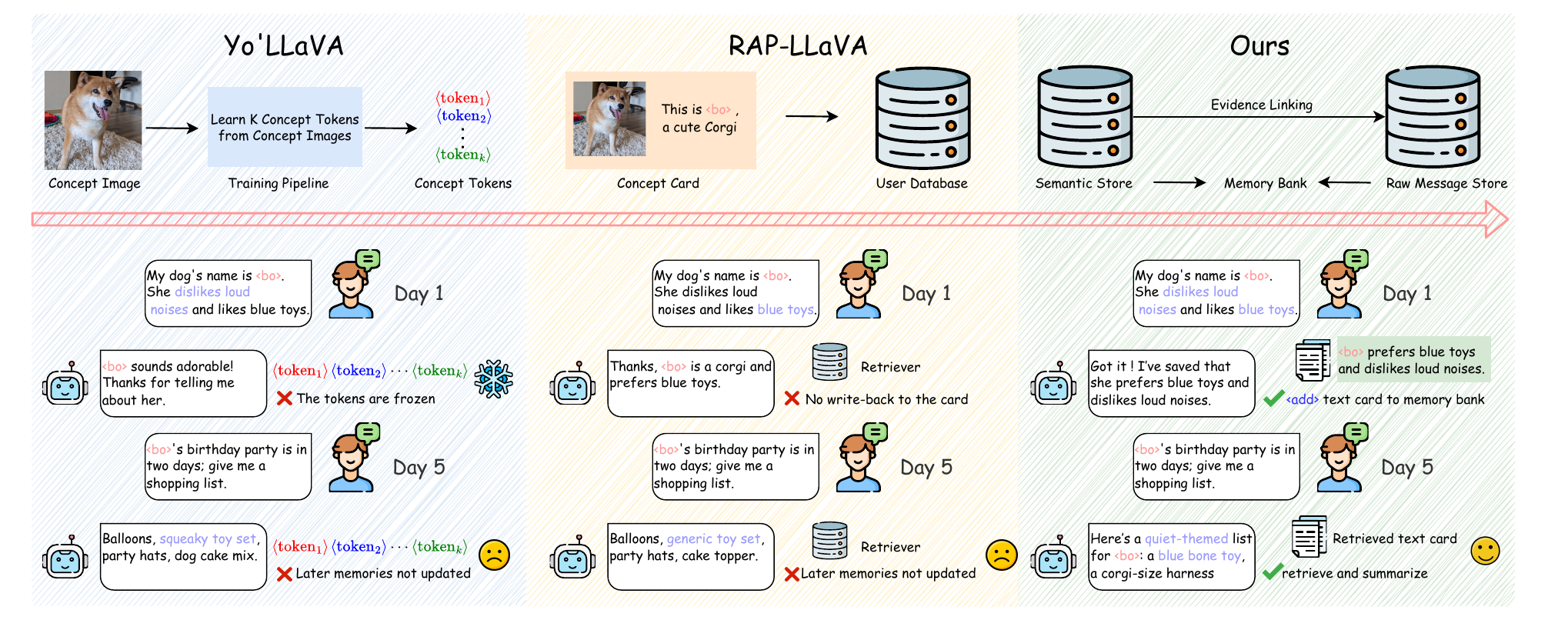}
   \caption{M$^{2}$A enables incremental personalization with an editable multimodal memory. Unlike Yo'LLaVA and RAP-LLaVA, which keep initial concept tokens or text only profiles without write-back, M$^{2}$A updates a unified memory bank during interaction and queries it at generation time, yielding recommendations aligned with evolving preferences across long, multi session dialogs.}
   \label{fig:demo}
\end{figure*}

We formalize long-term personalized interaction as a Partially Observable Markov Decision Process (POMDP)~\cite{pomdp}, where the user’s latent state $u$ evolves over time and is approximated by an explicit belief state $M_t$. Based on this formulation, we propose M$^{2}$A, an agentic framework with dual-layer hybrid memory. The system comprises two cooperating agents: (i) \textbf{ChatAgent}, which manages dialogue through a ReAct-style workflow (Query → Generate → Update) and decides when to access or modify memory; and (ii) \textbf{MemoryManager}, which holds exclusive read-write access and performs iterative, reasoning-driven retrieval and updates. Memory is organized into a lower \emph{Raw Message Store} that preserves complete conversational logs and an upper \emph{Semantic Memory Store} that maintains high-level semantic observations. Each semantic entry links to raw-message evidence via \texttt{evidence\_ids}, enabling evidence-linked progressive narrowing from coarse semantic retrieval to fine-grained context. A tri-path retrieval strategy combining dense text vectors, BM25~\cite{robert2009bm25} sparse retrieval, and cross-modal image embeddings with Reciprocal Rank Fusion~\cite{Cormack2009ReciprocalRF} further improves robustness.

Our contributions are summarized as follows:
\begin{enumerate}[leftmargin=*]
    \item \textbf{Agentic online personalized memory.} We introduce M$^{2}$A, an agentic multimodal memory framework that supports incremental concept updates during interaction.
    \item \textbf{Dual-layer hybrid memory with evidence linking.} We propose a two-tier memory architecture with progressive narrowing for efficient and precise retrieval.
    \item \textbf{Scalable multimodal data synthesis.} We design a pipeline that injects multimodal sub-sessions into long conversations for training and evaluation.
\end{enumerate}

\begin{figure*}[htb]
   \centering
   \includegraphics[width=\linewidth]{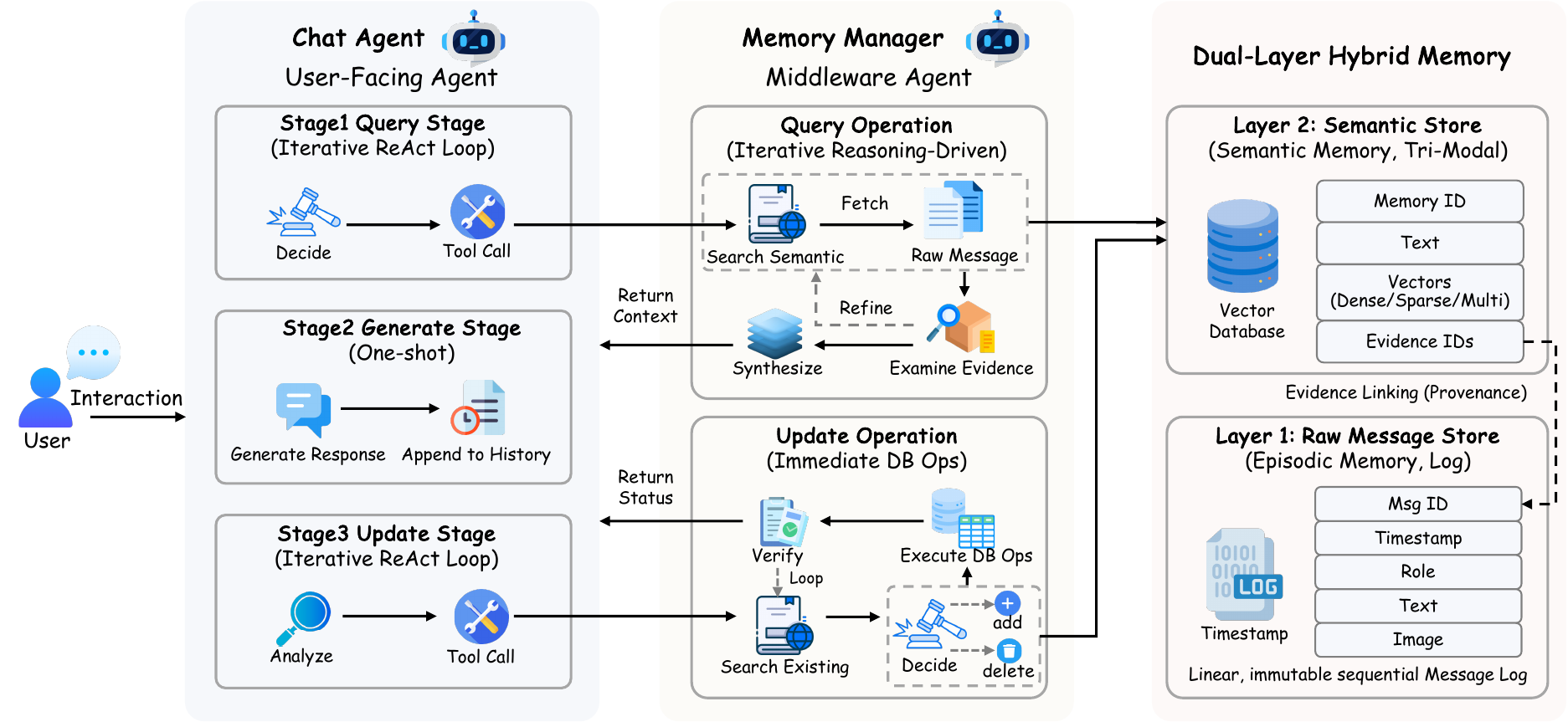}
   \caption{Overview of the $M^2A$ framework. $M^2A$ employs a multi-agent architecture consisting of a ChatAgent for user interaction and a MemoryManager for autonomous memory operations. The system leverages a Dual-Layer Hybrid Memory bank, linking high-level semantic observations in the Semantic Store to immutable conversational logs in the Raw Message Store via evidence IDs. }
   \label{fig:overview}
\end{figure*}

\section{Related Work}
\label{sec:related}

\paragraph{Personalized multimodal models.}
Recent vision-language models combine strong visual encoders with large language backbones~\cite{li2024llava_onevision,internvl2-2024,qwen2-vl-2024,llava-interleave-2024}, but largely overlook user-specific concepts. Existing personalization methods follow two paradigms. 

\textbf{Concept internalization} compresses user-provided concept images into embeddings injected into LLaVA-style models, enabling recognition of personalized visual entities at inference. Yo'LLaVA~\cite{nguyen2024yollava} targets single concepts, while MC-LLaVA~\cite{an2024mcllava} and MyVLM~\cite{alaluf2024myvlm} extend to multi-concept or few-shot settings. Yo'Chameleon~\cite{nguyen2025yoc} and UniCTokens~\cite{an2025unictokens} utilize this paradigm in the area of both personalized understanding and generation. Online-PVLM~\cite{bai2025onlinepvlm} introduces hyperbolic representations to enable online concept learning at test time without training, allowing incremental addition of new concepts during interaction. However, these approaches cannot flexibly update or refine existing concepts based on conversational feedback—once a concept is initialized (either through training or test-time encoding), its representation remains fixed. This limits their ability of online adaptation, refinement, or evolution of user's personalized knowledge.

\textbf{Retrieval-augmented personalization} instead stores user profiles, aliases, and concept descriptions externally and retrieves them during inference. RAP~\cite{hao2025rap} exemplifies this paradigm for multimodal language models, while Personalization Toolkit~\cite{seifi2025personalization} leverages pre-trained vision foundation models with visual prompting and retrieval to achieve training-free personalization. Such methods scale better by keeping the base model fixed, but typically assume a static concept set and lack mechanisms for online refinement of existing knowledge.

Our work builds on the retrieval-augmented paradigm while introducing \textbf{agentic memory updates}, allowing the system to autonomously decide when and how to modify personalized knowledge during interaction. Unlike concept internalization methods (including online learning variants) that encode concepts into fixed representations, or passive retrieval systems that merely store and retrieve static information, M2A maintains \textbf{editable, persistent memory} that accumulates, refines, and even corrects user-specific knowledge across multi-session conversations through explicit update operations.

\paragraph{Long-context memory and agentic memory management.}
As conversations exceed context windows, external memory becomes essential. LoCoMo shows that long dialogues require selectively reintroducing early utterances to maintain temporal consistency~\cite{maharana2024locomo}. Early systems adopt \textbf{passive retrieval}, storing conversation history externally and retrieving relevant segments at inference time (e.g., LongMem, MemLong, UniMem, HMT)~\cite{longmem-2023,memlong-2024}. While effective, these methods rely on fixed retrieval strategies.

\textbf{Agentic memory} introduces dynamic control over memory operations. MemGPT frames memory access as agent actions~\cite{memgpt-2023}, while MemoryBank, Mem0, and A-MEM extend this idea with long-term persistence and editable memories~\cite{zhong2024memorybank,amem-2025}. Despite their flexibility, existing systems are mostly text-based, rely on single-pass retrieval, and operate at a single memory granularity. We address these limitations with a \textbf{dual-layer hybrid memory} that supports multimodal concepts and \textbf{iterative reasoning-driven retrieval} from semantic summaries to raw conversational evidence.

Additional related work on multimodal generation and editing is discussed in Appendix~\ref{app:additional_related_work}.

\section{Problem Formulation}

We formalize long-term personalized interaction as a Partially Observable Markov Decision Process (POMDP), defined as the tuple $(\mathcal{U}, \mathcal{A}, \mathcal{T}, \Omega, \mathcal{O})$.

\paragraph{Latent user state.}
Let $\mathcal{U}$ denote the latent state space of user profiles. At each interaction round $t$, the user is in a hidden state $u_t \in \mathcal{U}$ that contains their private concepts (e.g., specific pet name ``Bobo''), visual preferences (e.g., ``likes blue toys''), and interaction styles evolving over time. The system cannot directly observe $u_t$ and can only infer it through user queries and feedback.

\paragraph{Transitions and observations.}
\begin{itemize}[leftmargin=*]
\item \textbf{State transition} $\mathcal{T}$: As the dialogue progresses, the user's latent state evolves according to transition probability $P(u_{t+1} | u_t, a_t)$. For example, when a user introduces new conceptual entities (``my dog is named Bobo'') or corrects previous preferences (``she actually prefers quiet toys'') during interaction, the latent state $u_t$ undergoes incremental updates.
\item \textbf{Observation} $\Omega$: At round $t$, the system receives a multimodal observation $x_t \in \Omega$ (the user's input query, potentially containing text and images). This observation is an instantiation of intent conditioned on the current latent state, following probability distribution $x_t \sim \mathcal{O}(x | u_t)$.
\end{itemize}

\paragraph{Belief state as memory bank.}
Due to the unobservability of $u_t$, the agent must maintain a \textbf{belief state} $M_t$ based on the historical observation sequence $H_t = \{x_1, a_1, \dots, x_t, a_t\}$. In our framework, $M_t$ is instantiated as a \textbf{multimodal memory bank}, whose goal is to approximate the posterior distribution of latent state $u_t$:
\begin{equation}
M_t \approx P(u_t | H_t)
\end{equation}

At each interaction round, the evolution mechanism of the memory bank is:
\begin{enumerate}[leftmargin=*]

\item \textbf{Action execution}: The system takes action $a_t \in \mathcal{A}$ (generates response) based on current observation $x_t$ and belief state $M_{t-1}$: $a_t \sim \pi(a | x_t, M_{t-1})$.
\item \textbf{Belief update}: The system updates memory through MemoryManager based on new observation $x_t$, system action $a_t$, and historical memory $M_{t-1}$: $M_t = f_{update}(M_{t-1}, x_t, a_t)$. 
\end{enumerate}

In M$^{2}$A, the belief state $M_t$ is instantiated through \textbf{dual-layer hybrid memory}: the lower RawMessageStore saves complete observation history $\{x_1, \dots, x_t\}$, while the upper SemanticAssociationStore stores high-level inferences about latent state $u_t$ (semantic observations). The collaboration mechanism between ChatAgent and MemoryManager implements the $f_{update}$ function: ChatAgent decides \textbf{when} to trigger updates, MemoryManager decides \textbf{how} to update (which memories to add, delete, or modify). The retrieval process corresponds to extracting information most relevant to the current query $x_t$ from $M_{t-1}$ to support action generation $a_t$ in a timely manner.

\begin{figure*}[htb]
   \centering
   \includegraphics[width=\linewidth]{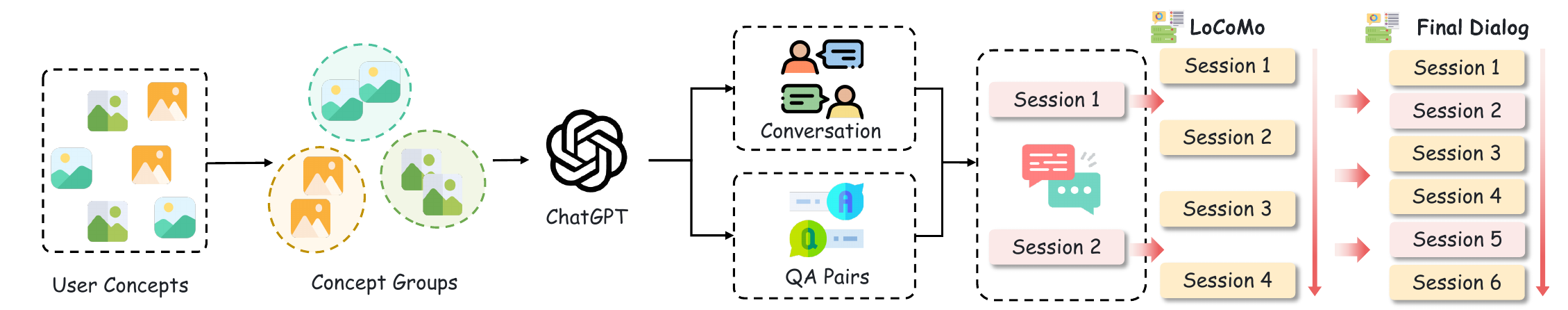}
   \caption{Overview of the proposed dataset construction pipeline. We first organize source images into semantic Concept Groups. Then, a unified One-Call Generation strategy produces concept-grounded dialogues and QA pairs. Finally, these generated sub-narratives are seamlessly interpolated into the original LoCoMo sessions to create the final hybrid dialog.}
   \label{fig:pipline}
\end{figure*}

\section{M$^{2}$A: Multimodal Memory Agent}

\subsection{Framework Overview}

Figure~\ref{fig:overview} illustrates M$^{2}$A's overall architecture. M$^{2}$A adopts a distributed collaboration architecture centered on MemoryManager, decoupling user interaction from memory management. The system consists of three main components:

\begin{enumerate}[leftmargin=*]
\item \textbf{ChatAgent (User Interaction Agent).} Serving as the system's front-end interface, ChatAgent conducts natural language conversations with users. The key feature is \textbf{autonomous decision-making capability}: in each conversational turn, ChatAgent autonomously determines whether to query or update long-term memory through a three-stage ReAct~\cite{yao2023react} workflow, rather than passively executing fixed procedures.

\item \textbf{MemoryManager (Memory Management Agent).} As the system's back-end core, MemoryManager is the only entity with read-write access to the memory bank. It executes \textbf{iterative reasoning-driven retrieval and update operations}: after receiving instructions from ChatAgent, it progressively narrows the retrieval scope through multi-round reasoning (from semantic memory to raw messages), or analyzes existing memories to decide how to update (create, delete, modify).

\item \textbf{Dual-Layer Hybrid Memory.} The system maintains two storage tiers: the lower \textbf{RawMessageStore} (raw message log) and the upper \textbf{SemanticMemoryStore} (semantic memory repository). The two layers are linked via \textbf{evidence\_ids}, enabling progressive narrowing from coarse to fine granularity.
\end{enumerate}

\subsection{Dual-Layer Hybrid Memory}

To address noise interference from long contexts and difficulties in cross-modal retrieval, we design a layered storage structure balancing retrospective accuracy with inference efficiency overall.

\paragraph{Layer 1: Raw Message Store.}
This is the memory's foundation, serving as an \textbf{append-only} database that stores raw conversational messages in chronological order. 

\paragraph{Layer 2: Semantic Memory Store.}
The upper layer stores high-level knowledge extracted and refined by MemoryManager. To address the challenge of images not being recalled by pure text queries, we introduce a \textbf{visual captioning} mechanism to incorporate visual content into searchable text. Each semantic entry $e$ can be represented as:
\begin{equation}
e = \{ c_{text}, c_{caption}, c_{image}, ptr \}
\end{equation}

where $c_{text}$ represents the textual semantic description (e.g., ``User's dog Bobo is a Corgi who likes blue toys''); $c_{caption}$ contains automatically generated captions for associated images using a vision-language model; $c_{image}$ is a image associated with this semantic entry; and $ptr = \{[s_1,e_1], [s_2,e_2], \ldots\}$ contains \texttt{evidence\_ids} linking to supporting raw message ranges in Layer 1. To speed up retrivial, we generate the following index vectors for each semantic entry: $v_{text}^{dense} \in \mathbb{R}^{d}$ is the dense text embedding computed from the concatenation of $c_{text}$ and $c_{caption}$ using a sentence encoder~\cite{reimers2019sentencebert} $E_T^{dense}$; $v_{text}^{sparse}$ denotes the BM25 sparse representation for keyword matching; $v_{img} \in \mathbb{R}^{d'}$ represents the visual embedding generated from associated images using a cross-modal encoder~\cite{zhai2023sigmoidlosslanguageimage} $E_I$.

\paragraph{Tri-path hybrid retrieval.}
Given user query $q$ (possibly text or text+image), retrieval score $S(q, e)$ is computed through three parallel paths:
\begin{align}
S_{dense}(q, e) &= \text{sim}(E_T^{dense}(q_{text}), v_{text}^{dense}) \\
S_{sparse}(q, e) &= \text{BM25}(q_{text}, v_{text}^{sparse}) \\
S_{visual}(q, e) &= \text{sim}(E_I(q_{img}), v_{img})
\end{align}

\textbf{Path 1 (Dense text semantic)}: Computes cosine similarity between the query's dense text embedding and each entry's $v_{text}^{dense}$, capturing semantic relationships.

\textbf{Path 2 (Sparse keyword)}: Applies BM25 scoring between query text and $v_{text}^{sparse}$, capturing exact keyword matches crucial for names, dates, and specific terms.

\textbf{Path 3 (Cross-modal)}: When $q_{img}$ is provided, computes similarity between the query image embedding and $v_{img}$; otherwise uses cross-modal text-to-image similarity between $q_{text}$ and $v_{img}$, enabling text queries to retrieve image-centric memories.

Finally, results from the three paths are fused via \textbf{Reciprocal Rank Fusion (RRF)}:
\begin{equation}
S_{RRF}(q, e) = \sum_{path \in \{dense, sparse, visual\}} \frac{1}{k + rank_{path}(e)}
\end{equation}
where $k=60$ is a hyperparameter and $rank_{path}(e)$ is the rank of entry $e$ in that path's results.

\textbf{Cross-modal alignment}: Since we generate $c_{caption}$ during storage, even when users search for images using only text ($q_{img}$ is empty), $q_{text}$ can recall corresponding image memories through high similarity with $c_{caption}$. This enables queries like ``my Corgi photo'' to return stored dog photos even if the semantic entry doesn't explicitly contain ``Corgi'' text description.

\subsection{Agentic Collaboration}
The system implements memory operations through the collaboration of two specialized agents, each with distinct responsibilities and capabilities.

\paragraph{ChatAgent workflow.}
ChatAgent processes each user message through three sequential stages inspired by ReAct:

\textbf{Query Stage}: Upon receiving a user message, ChatAgent first determines whether querying long-term memory is necessary. This decision considers whether the query references past events, people, or concepts not present in recent conversation context. If needed, ChatAgent queries memory, providing both the query and a snippet of recent conversation context. The agent may iterate this process, refining queries based on retrieved results, until sufficient information is gathered or a maximum iteration limit $N$ is reached. This iterative querying enables progressive information gathering and disambiguation in practice.

\textbf{Generate Stage}: With all the retrieved memory context and the current conversation history, ChatAgent generates a response to the user. The memory context is then naturally incorporated into the generation process through the language model's context window.

\textbf{Update Stage}: After generating the response, ChatAgent analyzes the conversation content to determine whether new information should be persisted. If updates are warranted, ChatAgent invokes MemoryManager to update memory.

\paragraph{MemoryManager operations.}
MemoryManager handles two types of requests from ChatAgent:

\textbf{Query operation}: Upon receiving a query request with accompanying conversation context, MemoryManager performs iterative reasoning-driven retrieval. It begins by searching the semantic memory bank using the tri-path hybrid retrieval mechanism. For promising candidates, it examines their \texttt{evidence\_ids} and may retrieve specific raw conversational segments for detailed context. This progressive narrowing from semantic to episodic memory enables MemoryManager to provide increasingly refined context. The agent may iterate this process (up to $N$ rounds), alternating between semantic search and raw message examination, until confident in having gathered sufficient relevant information for downstream answering.

\textbf{Update operation}: When receiving an update request, MemoryManager determines whether the current interaction introduces new, outdated, or conflicting information that should be reflected in long-term memory. Building on the same retrieval and reasoning mechanisms used in the query operation, it first inspects existing semantic memories related to the new content. Based on this comparison, MemoryManager may add new semantic entries, remove obsolete ones, or replace existing memories to maintain consistency. Throughout this process, the agent can re-query the memory store as needed to verify the effects of its actions and to avoid redundancy or contradiction. In essence, the update operation extends the query operation with write access to the semantic memory store, enabling dynamic and self-consistent memory maintenance.

\paragraph{Motivation for agent collaboration.}
The two-agent architecture provides several advantages over a single-agent design: First, it isolates memory management complexity from user interaction, preventing ChatAgent's conversation context from being overwhelmed by raw memory retrieval results. Second, MemoryManager can maintain its own reasoning context across operations without interfering with ChatAgent's dialogue context. Third, when memory operations require conversation context, ChatAgent provides only a concise recent snippet rather than full history, while MemoryManager can access older context through the memory bank itself when needed. This separation of concerns enables both agents to specialize in their respective tasks while collaborating effectively.

\subsection{Multimodal Chat Dataset Construction}
\label{subsec:dataset_construction}

Building upon the LoCoMo framework~\cite{maharana2024locomo}, we construct a concept-grounded multimodal chat corpus to address the limitation where images primarily serve as background context. As illustrated in \cref{fig:pipline}, our approach transforms visual inputs into essential narrative drivers through a four-stage pipeline:

\begin{enumerate}[leftmargin=*]
    \item \textbf{Concept Grouping:} Instead of isolated images, we sample semantically distinct \textit{Concept Groups} (e.g., specific objects or entities) to simulate realistic discussions.
    \item \textbf{Unified Multimodal Generation:} We employ a \enquote{One-Call} generation strategy using large language models. This method simultaneously generates multi-session dialogues and corresponding reasoning QA pairs, ensuring strict consistency between the visual evidence, the narrative flow, and the ground-truth answers.
    \item \textbf{Temporal Interpolation:} To maintain the temporal integrity of the original long-context conversations, generated sessions are inserted into specific time intervals of the host dialogue using strict timestamp interpolation.
    \item \textbf{Hybrid QA Injection:} The final evaluation set combines the newly generated reasoning questions (covering multi-hop, temporal, and open-domain types) with VQA samples explicitly injected into the dialogue stream.
\end{enumerate}

For comprehensive implementation details, including prompt constraints, QA category distributions, and hyperparameter settings, please refer to Appendix~\ref{sec:appendix_construction}.

\begin{table*}[ht]
\centering
\scriptsize
\setlength{\tabcolsep}{1.5pt} 
\renewcommand{\arraystretch}{1.2} 
\label{tab:main}
\resizebox{\textwidth}{!}{%
\begin{tabular}{c | c | ccc | ccc | ccc | ccc | ccc | ccc}
\hline
\multirow{2}{*}{\textbf{Model}} & \multirow{2}{*}{\textbf{Method}} 
& \multicolumn{3}{c|}{\textbf{Multi Hop}} 
& \multicolumn{3}{c|}{\textbf{Temporal}} 
& \multicolumn{3}{c|}{\textbf{Open Domain}} 
& \multicolumn{3}{c|}{\textbf{Single Hop}} 
& \multicolumn{3}{c|}{\textbf{Visual}} 
& \multicolumn{3}{c}{\textbf{Total}} \\ 

 &  & Q & G & Avg & Q & G & Avg & Q & G & Avg & Q & G & Avg & Q & G & Avg & Q & G & Avg \\
\hline

\multirow{4}{*}{\shortstack{GPT\\4o-mini}}
& LoCoMo          & 27.10 & 24.82 & 25.96 & 17.45 & 16.82 & 17.14 & \underline{36.14} & 29.17 & \underline{32.66} & 43.93 & 41.97 & 42.95 & 31.10 & 30.27 & 30.68 & 34.19 & 32.35 & 33.27 \\
& Mem0            & 25.89 & \underline{26.60} & 26.25 & 15.62 & 17.13 & 16.38 & 27.08 & 31.25 & 29.17 & 45.07 & 43.87 & 44.47 & \underline{36.39} & \underline{36.94} & \underline{36.67} & 34.66 & 34.80 & 34.73 \\
& A-MEM           & \textbf{28.36} & 25.18 & \underline{26.77} & \underline{24.61} & \underline{22.12} & \underline{23.37} & 30.21 & \underline{34.38} & 32.30 & \underline{45.18} & \underline{44.23} & \underline{44.71} & 36.39 & 36.70 & 36.54 & \underline{36.79} & \underline{35.73} & \underline{36.26} \\
\rowcolor{gray!30} & M$^{2}$A (ours) & \underline{27.30} & \textbf{28.01} & \textbf{27.66} & \textbf{31.15} & \textbf{32.40} & \textbf{31.78} & \textbf{40.63} & \textbf{36.46} & \textbf{38.55} & \textbf{56.24} & \textbf{56.71} & \textbf{56.48} & \textbf{44.03} & \textbf{42.51} & \textbf{43.27} & \textbf{44.61} & \textbf{44.67} & \textbf{44.64} \\
\hline

\multirow{4}{*}{\shortstack{Qwen3VL\\8b}}
& LoCoMo          & 24.82 & 23.76 & 24.29 & 28.97 & 29.28 & 29.13 & 31.25 & 30.21 & 30.73 & 43.88 & 48.04 & 45.96 & 37.31 & 35.17 & 36.24 & 36.64 & 37.98 & 37.31 \\
& Mem0            & 34.75 & 35.82 & 35.29 & \underline{33.64} & \underline{34.58} & \underline{34.11} & \underline{34.38} & \underline{33.33} & \underline{33.86} & \underline{54.34} & \underline{51.96} & \underline{53.15} & \underline{41.59} & \underline{39.14} & \underline{40.37} & \underline{44.56} & \underline{43.33} & \underline{43.95} \\
& A-MEM           & \underline{40.07} & \underline{40.78} & \underline{40.43} & 31.78 & 30.84 & 31.31 & 32.29 & 30.21 & 31.25 & 44.63 & 45.07 & 44.85 & 39.14 & 38.53 & 38.84 & 40.14 & 40.07 & 40.10 \\
\rowcolor{gray!30} &  M$^{2}$A (ours) & \textbf{43.26} & \textbf{43.62} & \textbf{43.44} & \textbf{39.25} & \textbf{42.67} & \textbf{40.96} & \textbf{51.04} & \textbf{48.96} & \textbf{50.00} & \textbf{66.71} & \textbf{63.14} & \textbf{64.93} & \textbf{54.13} & \textbf{51.68} & \textbf{52.90} & \textbf{55.44} & \textbf{53.94} & \textbf{54.69} \\
\hline

\multirow{4}{*}{\shortstack{GLM-4.6V\\Flash}}
& LoCoMo          & 25.18 & 24.47 & 24.83 & 24.61 & 22.12 & 23.37 & 32.29 & 27.08 & 29.69 & 45.18 & 43.76 & 44.47 & 35.17 & 32.11 & 33.64 & 36.21 & 34.23 & 35.22 \\
& Mem0            & \underline{43.97} & \underline{45.04} & \underline{44.51} & \underline{34.27} & \underline{33.02} & \underline{33.65} & \underline{39.58} & 35.42 & \underline{37.50} & \underline{57.91} & \underline{57.67} & \underline{57.79} & 40.37 & 39.45 & 39.91 & \underline{47.73} & \underline{47.19} & \underline{47.46} \\
& A-MEM           & 41.84 & 39.36 & 40.60 & 29.90 & 31.15 & 30.53 & 35.42 & \underline{36.46} & 35.94 & 49.94 & 45.90 & 47.92 & \underline{44.65} & \underline{41.90} & \underline{43.28} & 43.60 & 41.19 & 42.39 \\
\rowcolor{gray!30} & M$^{2}$A (ours) & \textbf{48.58} & \textbf{47.51} & \textbf{48.05} & \textbf{39.87} & \textbf{42.05} & \textbf{40.96} & \textbf{57.30} & \textbf{52.08} & \textbf{54.69} & \textbf{66.11} & \textbf{66.47} & \textbf{66.29} & \textbf{55.66} & \textbf{52.91} & \textbf{54.29} & \textbf{56.67} & \textbf{56.29} & \textbf{56.48} \\
\hline
\end{tabular}%
}
\caption{Experimental results on LoCoMo dataset. The best results are highlighted in \textbf{bold}, and the second-best results are \underline{underlined}. Models are evaluated on LLM-as-a-Judge with evaluation model Qwen3-VL-32B(Q), GPT-4o(G) and their average(Avg). }
\label{tab:main}
\end{table*}

\section{Experiment}

\subsection{Experimental Setup}

\paragraph{Datasets.}
We evaluate on our enhanced LoCoMo dataset constructed via the pipeline described in \Cref{subsec:dataset_construction}. The dataset consists of 10 long conversations averaging 621 turns and approximately 10k tokens each, with 214 images injected throughout the dialogues. The original LoCoMo questions are categorized into Single-Hop, Multi-Hop, Temporal, and Open Domain. Following prior work~\cite{yan2026memoryr1}, we exclude the Adversarial category due to missing ground-truth answers.

We further augment the benchmark with a new category, \textbf{Visual-Centric} questions, which are designed to fully leverage the injected visual information. These questions specifically test the model’s ability to recall and reason over visual content introduced during concept sessions. The Visual-Centric questions can be further divided into five subtypes: four of them are directly aligned with the original LoCoMo categories (Single-Hop, Multi-Hop, Temporal, and Open Domain) but require visual grounding, while the remaining subtype is adapted from the original QA pairs of YolLaVA. 

\paragraph{Baselines.}
We compare against three representative systems: \textbf{LoCoMo}~\cite{maharana2024locomo}: a RAG system with single-layer semantic vector storage and single-pass retrieval before generation. Below we will refer this system as RAG. \textbf{Mem0}~\cite{mem0}: A general memory layer supporting cross-session knowledge persistence but with text-only support. \textbf{A-MEM}~\cite{amem-2025}: An agentic memory system supporting editing operations but handling only text modality.

\paragraph{Evaluation metrics.}
Prior work typically uses F1 or BLEU-1 scores for evaluation. However, we argue these metrics are inadequate for open-ended generation tasks where answers may be semantically equivalent but lexically different (e.g., ``last week'' vs. specific dates). Instead, we employ \textbf{LLM-as-a-judge} evaluation using two independent judges: Qwen3-VL-32B~\cite{Bai2025Qwen3VL} (denoted as Q) and GPT-4o~\cite{OpenAI2024GPT4o} (denoted as G). Each judge assigns a binary correctness score by comparing generated answers to ground-truth references considering semantic equivalence. We report Q, G, and their average (Avg) as our primary metrics.

\paragraph{Implementation details.}
All local models are served using vLLM~\cite{kwon2023efficient} for efficient inference. For GPT-4o/GPT-4o-mini API calls, we use the official structured output API. Text embeddings for all methods use all-MiniLM-L6-v2~\cite{reimers2019sentencebert} (consistent with A-MEM). Image embeddings and cross-modal retrieval employ SigLIP-Base-Patch16-384~\cite{zhai2023sigmoidlosslanguageimage}. For the tri-path retrieval, we retrieve top-10 candidates from each path and fuse using RRF with $k=60$. Semantic memories are managed with Milvus~\cite{2021milvus}.

\subsection{Main Results}

\Cref{tab:main} reports experimental results across different question categories and base models. Overall, M$^{2}$A consistently outperforms all baselines across the majority of settings, demonstrating robust advantages in long-context and multimodal memory reasoning.

On GPT-4o-mini, M$^{2}$A achieves an average accuracy of 44.64\%, substantially surpassing RAG (33.27\%), Mem0 (34.73\%), and A-MEM (36.26\%). The gains are most pronounced on Single-Hop questions, where M$^{2}$A improves from 44.71\% (best baseline) to 56.48\%, highlighting the effectiveness of evidence-linked progressive narrowing for retrieving fine-grained and contextually relevant information. On the newly introduced Visual-Centric questions, M$^{2}$A reaches 43.27\% accuracy, significantly outperforming RAG (30.69\%), which validates the benefits of maintaining multimodal memory representations with image captions and cross-modal hybrid retrieval.

Similar trends are observed on open-source models. On Qwen3-VL-8B, M$^{2}$A achieves 54.69\% average accuracy, compared to 43.95\% for the strongest baseline. On GLM-4.6V-Flash, the performance gap narrows slightly (56.48\% vs. 47.46\%), yet M$^{2}$A consistently maintains a clear advantage across questions of all categories.

\begin{figure}[h]
   \centering
   \includegraphics[width=\columnwidth]{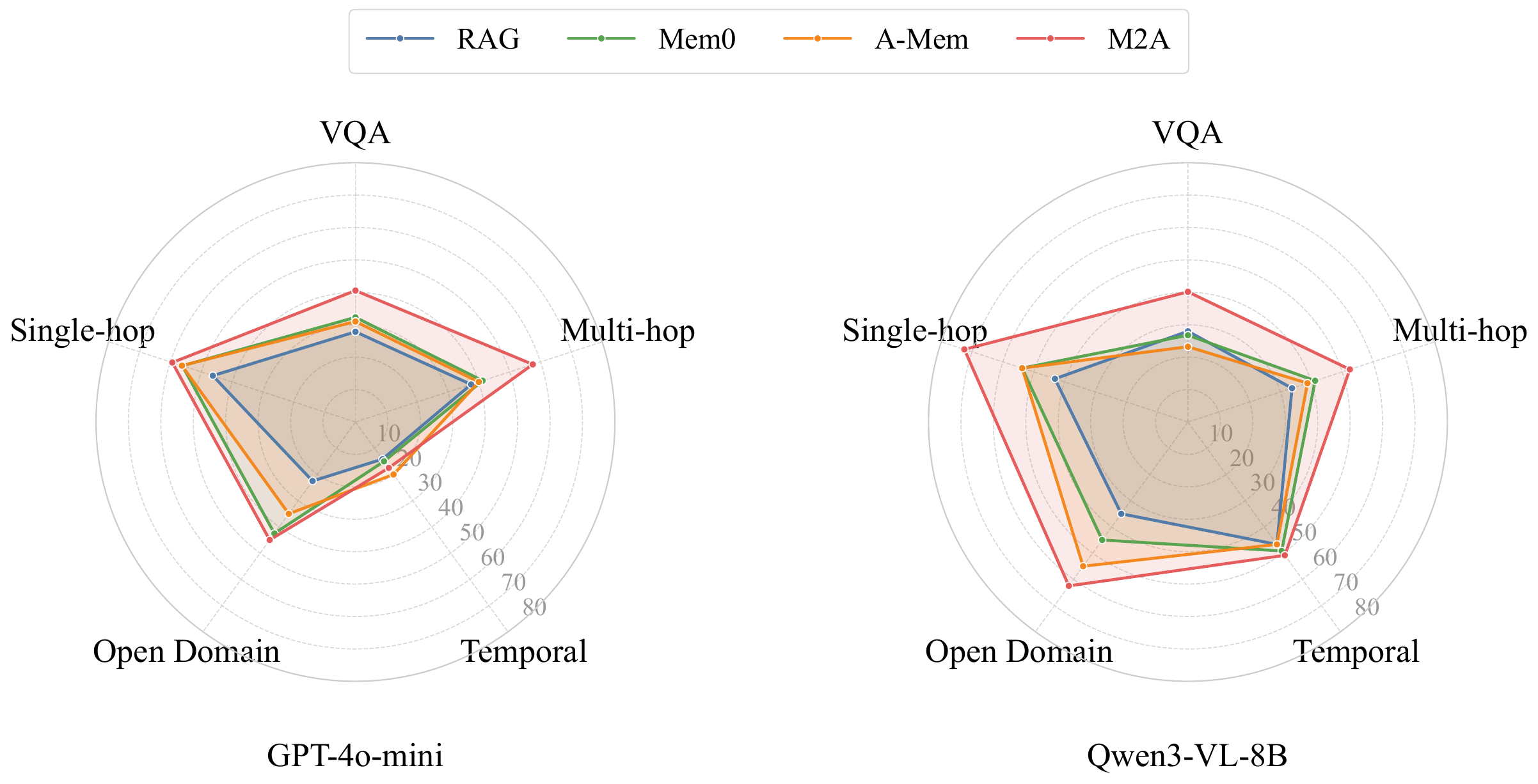}
   \caption{Detailed results on Visual-Centric questions.}
   \label{fig:visual}  
\end{figure}

\Cref{fig:visual} further presents a fine-grained breakdown of results on Visual-Centric questions for GPT-4o-mini and Qwen3-VL-8B. Across all subcategories, M$^{2}$A demonstrates superior performance, underscoring its strong capability in multimodal information recall and reasoning.

Notably, despite being an agentic memory system, A-MEM underperforms Pure RAG on several categories. We attribute this behavior to A-MEM’s text-only design and single-layer memory structure, which are insufficient for modeling multimodal concepts and supporting fine-grained temporal retrieval in our visually enriched benchmark.

\begin{table}[h]
\centering
\resizebox{\columnwidth}{!}{
\begin{tabular}{lccc}
\toprule
Variant & Qwen3-VL & GPT-4o & Avg. \\
\midrule
M$^{2}$A (Full) & 55.44 & 53.94 & 54.69 \\
w/o Dual-layer (semantic only) & 41.56 & 41.19 & 41.38 \\
w/o Iterative (single retrieval) & 38.08 & 39.26 & 38.67 \\
w/o Tri-path (text dense only) & 51.15 & 50.03 & 50.59 \\
\bottomrule
\end{tabular}%
}
\caption{Comparison of M$^{2}$A and its variation on Qwen3-VL-8B.}
\label{tab:ablation}
\end{table}

\subsection{Ablation Study}
To understand the contribution of each component, we conduct systematic ablations on the Qwen3VL-8B model. \cref{tab:ablation} presents results averaged across all question categories.

\paragraph{Effect of dual-layer memory.}
Removing the RawMessageStore and relying solely on SemanticMemoryStore (``w/o Dual-layer'') degrades performance by 13.31 percentage points. This validates our hypothesis that semantic summaries alone may lack crucial details or suffer from information loss during abstraction. The evidence-linked design enables MemoryManager to verify and augment semantic memories with precise raw context when needed.

\paragraph{Effect of iterative retrieval.}
Constraining MemoryManager to single-pass retrieval without iterative reasoning (``w/o Iterative'') reduces accuracy by 16.02 points. This demonstrates the value of progressive narrowing---the ability to examine initial semantic results, identify promising candidates through their linked raw messages, and iteratively refine the search. Single-pass retrieval often either misses relevant context or retrieves too broadly.

\paragraph{Effect of tri-path retrieval.}
Using only dense text embeddings without sparse BM25 or cross-modal image retrieval (``w/o Tri-path'') drops performance by 4.10 points. This results proves that the tri-path retrieval approach provides robustness: dense embeddings capture semantic similarity, BM25 handles exact entity matches, and cross-modal retrieval enables text-to-image queries. Removing any path weakens overall recall quality.

\begin{figure}[h]
   \centering
   \includegraphics[width=\columnwidth]{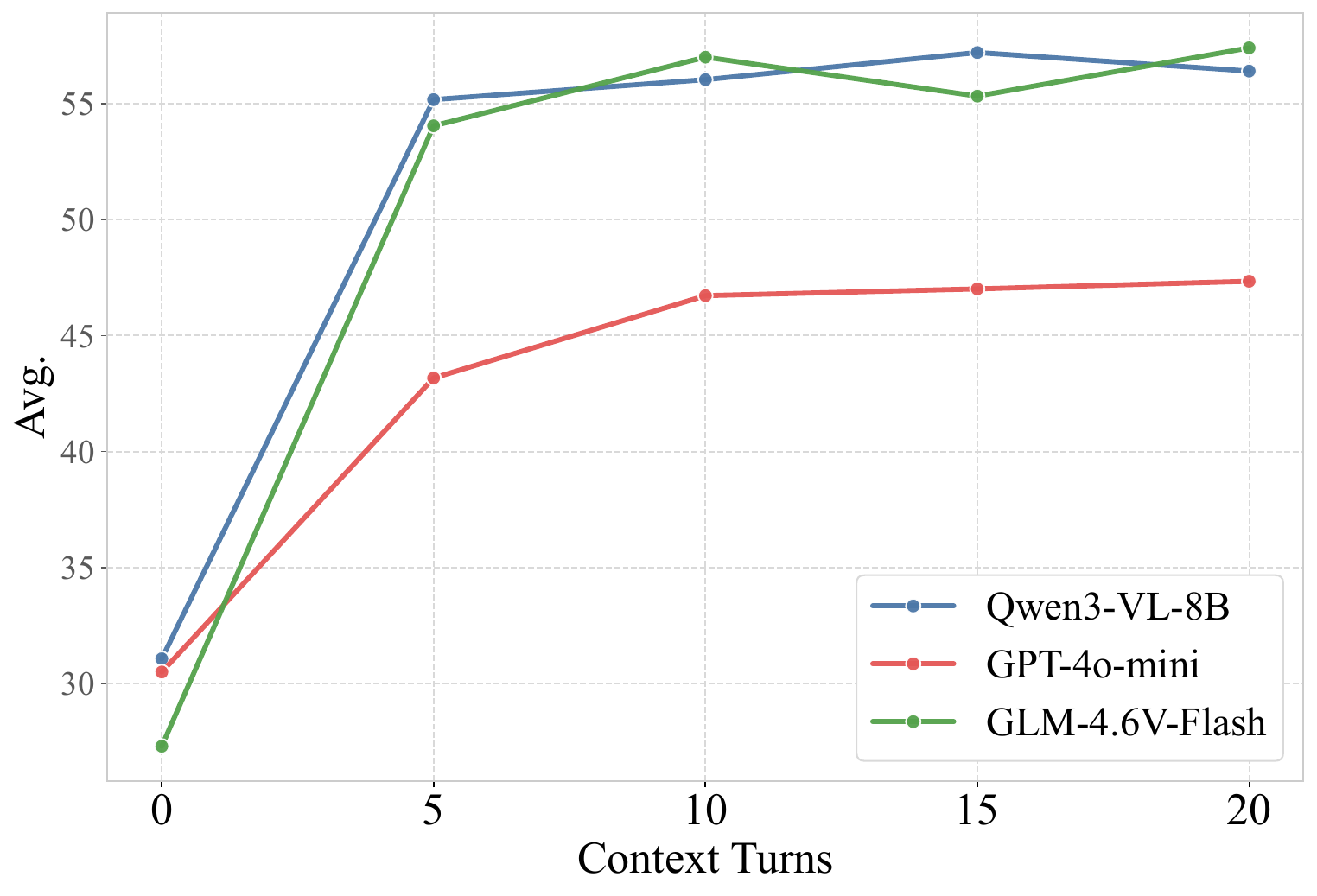}
   \caption{Detailed results on Visual-Centric questions.}
   \label{fig:ablation-context}  
\end{figure}

\paragraph{Effect of context window length.}
We study the impact of the amount of recent conversational context provided to MemoryManager when ChatAgent triggers memory queries or updates. Specifically, we vary the number of recent turns passed along with each request to the MemoryManager. As shown in \Cref{fig:ablation-context}, providing no contextual history severely degrades performance across all models, indicating that MemoryManager cannot reliably update or retrieve effective memories based solely on the current turn and ChatAgent’s high-level update suggestions.

Notably, supplying only a small amount of recent context (e.g., 5 turns) already leads to substantial performance gains, enabling MemoryManager to better infer user intent and anchor updates to the appropriate conversational state. Increasing the context window further yields only marginal improvements. Overall, these results suggest that a short recent context window is sufficient for effective memory operations, while longer contexts provide limited additional benefits and primarily incur extra computational overhead.

\section{Conclusion}

We presented M$^{2}$A, an agentic multimodal memory system addressing the challenge of incremental personalization in long-term human-machine interactions. Through dual-layer hybrid memory with evidence linking, M$^{2}$A enables progressive narrowing from semantic summaries to fine-grained conversational context. The collaboration between ChatAgent and MemoryManager implements autonomous, reasoning-driven memory operations that adapt dynamically to conversation context. Tri-path cross-modal hybrid retrieval combining dense text embeddings, BM25 sparse retrieval, and cross-modal image embeddings provides robust recall across semantic, lexical, and visual cues. Experimental results demonstrate substantial improvements over existing methods, particularly on temporally complex and visual-centric questions. This work shows that transforming personalization from static configuration to co-evolving memory provides a viable path toward more adaptive and individualized AI systems.

\section*{Impact Statement}

This paper presents work advancing long-term personalized interaction in AI systems. There are many potential societal consequences of our work, none of which we feel must be specifically highlighted here.

\nocite{langley00}

\bibliography{example_paper}
\bibliographystyle{icml2026}

\newpage
\appendix
\crefalias{section}{appendix}
\crefalias{subsection}{appendix}
\onecolumn
\section{Additional Related Work}
\label{app:additional_related_work}

Beyond memory-centric personalization approaches, there exists a complementary research direction on controllable multimodal generation and editing systems. While our work focuses on personalization through memory and retrieval, these systems explore personalization through conditional generation, offering an alternative perspective on user-specific multimodal systems.

\paragraph{Controllable generation with user constraints.}
Diffusion-based generative models have enabled fine-grained control over visual synthesis under user-specified constraints. \citet{zeng2024cat} propose Cat-DM, a controllable accelerated virtual try-on system that leverages diffusion models to render garments according to user-defined constraints while maintaining synthesis efficiency. This demonstrates how generative priors can be conditioned on user inputs to produce personalized visual outputs. Such conditional generation paradigms represent an orthogonal approach to personalization: rather than retrieving and reasoning over stored user information, these systems directly encode user preferences into generation controls.

\paragraph{Multimodal editing and enhancement.}
Image editing techniques that integrate multimodal signals provide another avenue for user-driven customization. \citet{song2023fashion} explore fashion customization through image generation based on editing cues, enabling users to modify garment appearance and style through natural specifications. Building on this direction, \citet{song2025mef} introduce MEF-GD, a multimodal enhancement and fusion network for garment design that integrates diverse modalities to improve synthesis quality and support richer semantic controls. These works highlight the role of multimodal fusion in enabling more expressive user control over generated content.

\paragraph{Cross-view consistency and alignment.}
Maintaining consistency across different views or poses presents additional challenges for personalized visual systems. \citet{zhang2025robust} present Robust-MVTON, which learns cross-pose feature alignment and fusion to achieve robust multi-view virtual try-on results across varied user poses and viewpoints. This work emphasizes the importance of robust feature alignment when personalizing visual content under geometric variations.

While these generation-based systems differ from our memory-centric approach, both paradigms ultimately aim to create user-specific experiences in multimodal contexts. Generation-based methods encode personalization directly into model outputs through conditional controls, whereas memory-based methods (including ours) maintain explicit user representations that guide response generation through retrieval. These approaches can be viewed as complementary: generation-based systems excel at producing novel personalized visual content, while memory-based systems excel at maintaining consistency and reasoning over accumulated user knowledge across extended interactions. Future work might explore hybrid approaches that combine editable long-term memory with controllable generation capabilities.

\section{Detailed Baselines Introduction}

\subsection{A-MEM}
We use the A-MEM baseline. The system initializes the \texttt{AgenticMemorySystem} from the original A-MEM codebase with a sentence embedding model (\texttt{all-MiniLM-L6-v2}) and connects the LLM to a local vLLM endpoint via the API. Each conversation turn is converted into a structured memory note that includes the timestamp, speaker name, and message content. If images are present, the system first performs image captioning with the same vision-language model and appends a concise natural-language image summary to the memory note. During question answering, it retrieves top-$k$ related memories (default $k=10$) and constructs the final prompt from the retrieved context. This implements an agentic memory read/write pipeline while keeping the base model fixed.

\subsection{Mem0}
We use the legacy Mem0 implementation with an in-memory Qdrant vector store and a embedding model (\texttt{all-MiniLM-L6-v2}). Similar to A-MEM, each dialog turn is ingested into memory with optional image captioning; captions are appended to the text as a system note so that vision signals are converted into searchable text. At inference, it retrieves a fixed number of memories (default $k=10$), builds a retrieval-augmented prompt, and generates the answer with the vLLM-backed model.

\subsection{RAG}
This is a single-pass retrieval setup without explicit memory editing. All conversation turns are embedded with a sentence-transformer model and stored as dense vectors. At query time, we compute cosine similarity (dot product on normalized embeddings) between the question and all stored turns, select the top-$k$ contexts (default $k=5$), and concatenate them into the prompt. Image paths from retrieved turns are passed through the model as additional visual inputs. This provides a pure retrieval-augmented baseline without long-term memory updates.


\section{Dataset Construction Details}
\label{sec:appendix_construction}

This section details the pipeline implemented to construct the concept-grounded multimodal dataset. The process is governed by strict constraints to ensure high-quality visual grounding and temporal coherence.

\paragraph{Step 1: Concept Grouping and Sampling.}
For each target conversation, we sample a concept group $\mathcal{G} = \{c_1, ..., c_k\}$ where the group size $k$ is uniformly sampled from $[3, 4]$. For each concept $c_i$, we retrieve a subset of $m \in [2, 3]$ images from the source dataset to ensure visual consistency across different dialogue turns.

\paragraph{Step 2: One-Call Generation Strategy.}
We utilize GPT-4 to generate the dialogue and QA pairs in a single API call to maximize coherence. The generation is constrained by the following hyperparameters:
\begin{itemize}
    \item \textbf{Session Structure:} The model generates $N_{sess} \in [5, 6]$ distinct sessions. Each session consists of $T_{turns} \in [5, 15]$ dialogue turns.
    \item \textbf{Visual Grounding:} The first message of the sequence is forced to explicitly reference all concepts in $\mathcal{G}$ using angle-bracket notation (e.g., \texttt{<concept\_name>}) to establish the entities.
\end{itemize}

\paragraph{Step 3: QA Taxonomy and Distribution.}
To balance the difficulty of the benchmark, we enforce a strict distribution ratio of \textbf{2:3:1:4} for the generated QA categories. For a typical set of 20 generated questions, the distribution is:
\begin{itemize}
    \item \textbf{Type 6 (Multi-Hop):} 20\% (4 questions). Requires reasoning across multiple dialogue turns.
    \item \textbf{Type 7 (Temporal):} 30\% (6 questions). Involves understanding the timeline of events.
    \item \textbf{Type 8 (Open-Domain):} 10\% (2 questions). Requires external knowledge grounded in the image.
    \item \textbf{Type 9 (Single-Hop):} 40\% (8 questions). Direct information retrieval.
\end{itemize}
Additionally, we inject \textbf{Type 5 (Visual Retrieval)} questions by matching images appearing in the dialogue with existing VQA datasets.

\paragraph{Step 4: Temporal Interpolation.}
To insert the generated sessions into a host conversation without disrupting its timeline, we perform linear time interpolation. Let the insertion point in the host conversation be between timestamps $t_{start}$ and $t_{end}$. We generate timestamps $\tau_j$ for the $j$-th generated session ($j \in \{1, ..., N_{sess}\}$) using:
\begin{equation}
    \tau_j = t_{start} + \frac{j}{N_{sess} + 1} \times (t_{end} - t_{start})
\end{equation}
This ensures that $t_{start} < \tau_1 < ... < \tau_{N_{sess}} < t_{end}$, preserving the strictly increasing temporal order of the merged conversation.

\section{System Prompts}
\label{app:prompts}

\subsection{ChatAgent Prompts}
\begin{systemprompt}[ChatAgent System Prompt]

You are a helpful AI assistant with access to long-term memory. Your goal is to provide personalized, contextually appropriate responses based on the user's conversation history.

\vspace{0.8em}
\textcolor{accentpurple}{\faTools\hspace{0.5em}\textbf{Available Tools:}}
\vspace{0.3em}

\begin{itemize}[leftmargin=1.2em, itemsep=2pt, label=\textcolor{accentgreen}{\faCube}]
  \item \tool{query\_memory} \textcolor{commentcolor}{— Search long-term memory for relevant information}
  \item \tool{update\_memory} \textcolor{commentcolor}{— Request memory updates for important information}
\end{itemize}

\vspace{0.8em}
\textcolor{accentblue}{\hspace{0.5em}\textbf{Workflow:}}
\vspace{0.3em}

Each user message goes through three stages:

\vspace{0.5em}
\begin{tcolorbox}[
  colback=codebg,
  colframe=codebg,
  arc=3pt,
  boxrule=0pt,
  left=8pt, right=8pt, top=6pt, bottom=6pt,
]
\stage{\faSearch\hspace{0.3em}Stage 1: Query}\\[4pt]
\textcolor{prompttext}{Decide if you need information from long-term memory. Query when:}
\begin{itemize}[leftmargin=1.2em, itemsep=1pt, label=\textcolor{accentblue}{\faChevronRight}, topsep=3pt]
  \item User asks about past events, people, or facts not in recent context
  \item User references concepts introduced long ago
  \item Answering requires temporal reasoning across sessions
\end{itemize}
\textcolor{commentcolor}{\footnotesize You may query multiple times iteratively. Use retrieved information to refine subsequent queries if needed.}
\end{tcolorbox}

\vspace{0.3em}
\begin{tcolorbox}[
  colback=codebg,
  colframe=codebg,
  arc=3pt,
  boxrule=0pt,
  left=8pt, right=8pt, top=6pt, bottom=6pt,
]
\stage{\faCommentDots\hspace{0.3em}Stage 2: Generate}\\[4pt]
\textcolor{prompttext}{With all retrieved memory + recent conversation, generate a natural, helpful response.}
\end{tcolorbox}

\vspace{0.3em}
\begin{tcolorbox}[
  colback=codebg,
  colframe=codebg,
  arc=3pt,
  boxrule=0pt,
  left=8pt, right=8pt, top=6pt, bottom=6pt,
]
\stage{\faSave\hspace{0.3em}Stage 3: Update}\\[4pt]
\textcolor{prompttext}{Decide if the conversation contains information worth remembering:}
\begin{itemize}[leftmargin=1.2em, itemsep=1pt, label=\textcolor{accentgreen}{\faChevronRight}, topsep=3pt]
  \item Personal information \textcolor{commentcolor}{(events, plans, preferences)}
  \item Factual details \textcolor{commentcolor}{(names, relationships, attributes)}
  \item Concept introductions or refinements
  \item Important milestones or changes
\end{itemize}
\textcolor{prompttext}{Call \tool{update\_memory} if warranted.}
\end{tcolorbox}

\vspace{0.8em}
\textcolor{accentred}{\faExclamationTriangle\hspace{0.5em}\textbf{Important:}}
\vspace{0.3em}

\begin{itemize}[leftmargin=1.2em, itemsep=3pt, label=\textcolor{accentred}{\faCheckCircle}]
  \item \important{Be selective} about memory operations — not every turn needs querying or updating
  \item When querying/updating, provide \keyword{clear, specific requests}
  \item Recent conversation context is provided automatically
\end{itemize}

\end{systemprompt}

\subsection{MemoryManager Prompts}
\begin{systemprompt}[MemoryManager System Prompt]

You are the memory management agent responsible for all read/write operations 
on the memory bank. Your role is to perform precise, reasoning-driven retrieval 
and updates.

\vspace{0.8em}
\textcolor{accentpurple}{\faTools\hspace{0.5em}\textbf{Available Tools:}}
\vspace{0.3em}

\begin{itemize}[leftmargin=1.2em, itemsep=2pt, label=\textcolor{accentgreen}{\faCube}]
  \item \tool{search\_semantic\_memories} \textcolor{commentcolor}{— Search semantic memory using tri-path retrieval}
  \item \tool{fetch\_raw\_messages} \textcolor{commentcolor}{— Retrieve raw messages by ID ranges}
  \item \tool{add\_memory} \textcolor{commentcolor}{— Create new semantic memory entry}
  \item \tool{delete\_memory} \textcolor{commentcolor}{— Delete semantic memory entry}
\end{itemize}

\vspace{0.8em}
\textcolor{accentblue}{\faSearch\hspace{0.5em}\textbf{Query Operation Workflow:}}
\vspace{0.3em}

When receiving a query request with conversation context:
\begin{enumerate}[leftmargin=1.5em, itemsep=2pt, label=\textcolor{accentblue}{\arabic*.}]
  \item Search semantic memories using \keyword{tri-path retrieval}
  \item Examine promising candidates' \keyword{evidence\_ids}
  \item If needed, fetch corresponding raw messages for details
  \item Iteratively refine: \textcolor{accentgreen}{search} → \textcolor{accentyellow}{examine} → \textcolor{accentpurple}{fetch}
  \item Synthesize and return refined memory context
\end{enumerate}

\vspace{0.5em}
\begin{tcolorbox}[
  colback=codebg,
  colframe=codebg,
  arc=3pt,
  boxrule=0pt,
  left=8pt, right=8pt, top=6pt, bottom=6pt,
]
\textcolor{accentyellow}{\faLightbulb}\hspace{0.5em}\textcolor{prompttext}{\textbf{Progressive Narrowing Strategy:}}\\[3pt]
\textcolor{commentcolor}{
  Start broad → Identify clusters → Drill down via evidence\_ids → Repeat
}
\end{tcolorbox}

\vspace{0.8em}
\textcolor{accentgreen}{\faEdit\hspace{0.5em}\textbf{Update Operation Workflow:}}
\vspace{0.3em}

When receiving update request with conversation context:
\begin{enumerate}[leftmargin=1.5em, itemsep=2pt, label=\textcolor{accentgreen}{\arabic*.}]
  \item Search for related existing memories
  \item Decide strategy: \keyword{CREATE} / \keyword{DELETE} / \keyword{BOTH}
  \item If deleting, also add update record describing the change
  \item Execute operations with appropriate \keyword{evidence\_ids}
  \item May query again between operations for verification
\end{enumerate}

\vspace{0.8em}
\textcolor{accentred}{\faExclamationTriangle\hspace{0.5em}\textbf{Important Principles:}}
\vspace{0.3em}

\begin{itemize}[leftmargin=1.2em, itemsep=3pt, label=\textcolor{accentred}{\faCheckCircle}]
  \item \important{Avoid duplicate memories} — check for existing similar entries
  \item Link semantic memories to supporting raw messages via \keyword{evidence\_ids}
  \item When updating contradictory info: \textcolor{accentred}{delete old} → \textcolor{accentgreen}{create new} → add update record
  \item Be precise with evidence\_ids — reference exact relevant message ranges
\end{itemize}

\end{systemprompt}

\newpage
\subsection{LLM-as-a-Judge Prompts}
We adapt the same LLM-as-a-Judge Prompts from Memory-R1~\cite{yan2026memoryr1}:

\begin{systemprompt}[Evaluation Judge Prompt]

Your task is to label an answer to a question as \correct{CORRECT} or \wrong{WRONG}.

\vspace{0.8em}
\textcolor{accentpurple}{\faDatabase\hspace{0.5em}\textbf{Input Data:}}
\vspace{0.3em}

\begin{enumerate}[leftmargin=1.5em, itemsep=2pt, label=\textcolor{accentpurple}{(\arabic*)}]
  \item A \keyword{question} \textcolor{commentcolor}{— posed by one user to another user}
  \item A \keyword{gold answer} \textcolor{commentcolor}{— ground truth answer}
  \item A \keyword{generated answer} \textcolor{commentcolor}{— answer to be evaluated}
\end{enumerate}

\vspace{0.5em}
\textcolor{commentcolor}{The point of the question is to ask about something one user should know about the other user based on their prior conversations.}

\vspace{0.8em}
\textcolor{accentblue}{\faBalanceScale\hspace{0.5em}\textbf{Grading Guidelines:}}
\vspace{0.3em}

\begin{tcolorbox}[
  colback=codebg,
  colframe=codebg,
  arc=3pt,
  boxrule=0pt,
  left=8pt, right=8pt, top=6pt, bottom=6pt,
]
\textcolor{accentorange}{\faComments\hspace{0.3em}General Questions}\\[4pt]
\textcolor{prompttext}{The gold answer is usually concise. The generated answer might be longer, but \keyword{be generous} — as long as it touches on the same topic, mark it \correct{CORRECT}.}\\[6pt]
\textcolor{commentcolor}{\faLightbulb\hspace{0.3em}Example:}\\
\textcolor{prompttext}{Q: Do you remember what I got the last time I went to Hawaii?}\\
\textcolor{prompttext}{Gold: \textit{A shell necklace}}
\end{tcolorbox}

\vspace{0.3em}
\begin{tcolorbox}[
  colback=codebg,
  colframe=codebg,
  arc=3pt,
  boxrule=0pt,
  left=8pt, right=8pt, top=6pt, bottom=6pt,
]
\textcolor{accentorange}{\hspace{0.3em}Time-Related Questions}\\[4pt]
\textcolor{prompttext}{Gold answer will be a specific date/month/year. Generated answer might use relative references (e.g., "last Tuesday"). \keyword{Be generous} — if it refers to the same time period, mark \correct{CORRECT} even if format differs.}\\[6pt]
\textcolor{commentcolor}{\faLightbulb\hspace{0.3em}Example: "May 7th" $\approx$ "7 May" → \correct{CORRECT}}
\end{tcolorbox}

\vspace{0.8em}
\textcolor{accentyellow}{\faQuestionCircle\hspace{0.5em}\textbf{Evaluation Input:}}
\vspace{0.3em}

\begin{itemize}[leftmargin=1.2em, itemsep=3pt, label=\textcolor{accentteal}{\faChevronRight}]
  \item \textbf{Question:} \variable{question}
  \item \textbf{Gold answer:} \variable{gold\_answer}
  \item \textbf{Generated answer:} \variable{generated\_answer}
\end{itemize}

\vspace{0.8em}
\textcolor{accentgreen}{\faFileExport\hspace{0.5em}\textbf{Output Format:}}
\vspace{0.3em}

\begin{enumerate}[leftmargin=1.5em, itemsep=2pt, label=\textcolor{accentgreen}{\arabic*.}]
  \item Provide a \keyword{short (one sentence)} explanation of your reasoning
  \item Finish with \correct{CORRECT} or \wrong{WRONG}
  \item Return in JSON format with key \textcolor{accentteal}{"label"}
\end{enumerate}

\vspace{0.5em}
\begin{tcolorbox}[
  colback=codebg,
  colframe=accentred,
  arc=3pt,
  boxrule=1pt,
  left=8pt, right=8pt, top=6pt, bottom=6pt,
]
\textcolor{accentred}{\faExclamationCircle\hspace{0.3em}\textbf{Warning:}} Do \wrong{NOT} include both CORRECT and WRONG in your response, or it will break the evaluation script.
\end{tcolorbox}

\end{systemprompt}

\end{document}